\documentclass[journal]{IEEEtran}
\ifCLASSINFOpdf
   \usepackage[pdftex]{graphicx}
   \DeclareGraphicsExtensions{.pdf,.jpeg,.png}
\else
\fi
\ifCLASSOPTIONcompsoc
 \usepackage[caption=false,font=normalsize,labelfont=sf,textfont=sf]{subfig}
\else
\usepackage[caption=false,font=footnotesize]{subfig}
\fi

\usepackage[dvipsnames]{xcolor}
\begin{document}
%
\title{Integrating Optimization Theory with Deep Learning for Wireless Network Design }
%
%
%

\author{Sinem~Coleri,~\IEEEmembership{Fellow,~IEEE}, Aysun Gurur Onalan,~\IEEEmembership{Student Member,~IEEE} and Marco~di Renzo,~\IEEEmembership{Fellow,~IEEE} \thanks{S. Coleri and A. G. Onalan are with the Department
of Electrical and Electronics Engineering, Koc University, Istanbul, e-mail: \{scoleri, aonalan17\}@ku.edu.tr. M. Di Renzo is with Universit\'e Paris-Saclay, CNRS, CentraleSup\'elec, Laboratoire des Signaux et Syst\`emes, 3 Rue Joliot-Curie, 91192 Gif-sur-Yvette, France, email: marco.di-renzo@universite-paris-saclay.fr. \\ \indent
Sinem Coleri acknowledges the support of the Scientific and Technological
Research Council of Turkey 2247-A National Leaders Research Grant
\#121C314. The work of M. Di Renzo was supported in part by the European
Commission through the Horizon Europe project COVER under grant
agreement number 101086228, the Horizon Europe project UNITE under
grant agreement number 101129618, and the Horizon Europe project
INSTINCT under grant agreement number 101139161, as well as by the
Agence Nationale de la Recherche (ANR) through the France 2030 project
ANR-PEPR Networks of the Future under grant agreement NF-Founds 22-PEFT-0010, and by the CHIST-ERA project PASSIONATE under grant
agreements CHIST-ERA-22-WAI-04 and ANR-23-CHR4-0003-01.
}}

%
%

\markboth{SUBMITTED TO IEEE COMMUNICATIONS MAGAZINE}%
{Shell \MakeLowercase{\textit{et al.}}: Bare Demo of IEEEtran.cls for IEEE Journals}
%



\maketitle

\begin{abstract}
Traditional wireless network design relies on optimization algorithms derived from domain-specific mathematical models, which are often inefficient and unsuitable for dynamic, real-time applications due to high complexity. Deep learning has emerged as a promising alternative to overcome complexity and adaptability concerns, but it faces challenges such as accuracy issues, delays, and limited interpretability due to its inherent black-box nature.  This paper introduces a novel approach that integrates optimization theory with deep learning methodologies to address these issues. The methodology starts by constructing the block diagram of the optimization theory-based solution, identifying key building blocks corresponding to optimality conditions and iterative solutions. Selected building blocks are then replaced with deep neural networks, enhancing the adaptability and interpretability of the system. Extensive simulations show that this hybrid approach not only reduces runtime compared to optimization theory based approaches but also significantly improves accuracy and convergence rates, outperforming pure deep learning models.

\end{abstract}


%
\IEEEpeerreviewmaketitle

\section{Introduction}

Conventional wireless network design relies on optimization theory-based algorithms to enhance system performance while adhering to various constraints imposed by regulations, applications, and communication feasibility (e.g., \cite{Onalan20, opt_3}). Formulating these optimization problems requires restrictive assumptions to model system performance and constraints, typically based on the statistical analysis of experimental data and domain knowledge. The complex interconnections between network layers necessitate a cross-layer design, making these problems difficult to solve. While iterative or heuristic algorithms are often used, they still have high time complexity, limiting their adaptability to real-time changes in large-scale networks.

To tackle the intricate complexity and potential intractability of problem formulations and enhance adaptability in learning near-real-time communication characteristics, recent solutions have explored deep learning methods (e.g., \cite{Onalan22, Khan22, opt_dnn_03}). In deep learning, the system parameters are mapped to optimal resource allocation based on the collected data over time. Training the deep learning architecture by using real data ensures adaptability while significantly reducing the complexity of the solutions. This approach also effectively addresses inaccuracies present in conventional model-based approaches due to simplifying assumptions or evolving scenarios. However, this approach may still face accuracy challenges, particularly concerning convergence time and the satisfaction of constraints due to the approximation inherent in deep learning, and frequent inclusion of regularizing
terms and mechanisms such as dropout and batch normalization. Moreover, deep learning approach demands a vast amount of data for training, leading to substantial communication overhead and significant delays. Furthermore, the resulting models often lack interpretability due to their black-box nature, casting doubt on their applicability across diverse scenarios.
Combining model-based and deep learning solutions holds the promise of mitigating these challenges.

 \begin{figure*}[!t]
\centering
\includegraphics[width=5.0in]{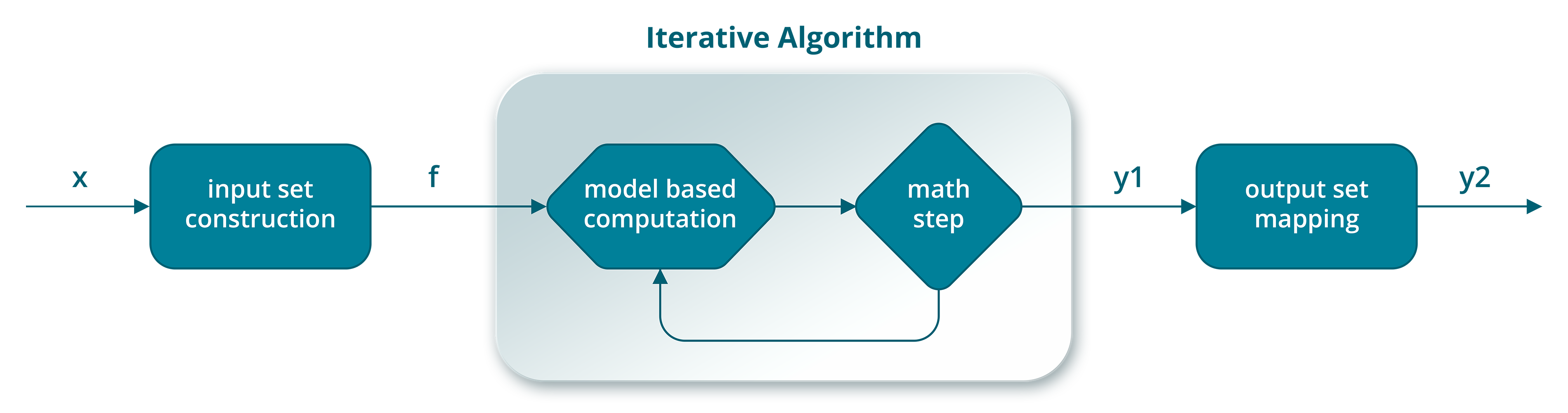}
\caption{Optimization theory based approach.}
\label{fig_opt}
\end{figure*}

In the transition from optimization theory to deep learning, several innovative approaches have emerged to streamline the training process. One set of these methodologies aims to reduce the amount of online training data by pretraining them with data generated using approximate models. These approximate models can be either optimization theory-based or simulation-based, as applied in learning-to-optimize frameworks \cite{learn_to_optimize} and knowledge assisted training architectures like digital twins \cite{digital_twin}. 
However, these methods often treat the mapping between system parameters and optimal resource allocation as a black box, without delving into the inner workings of the model-based solutions. Another set of the innovative approaches focuses on splitting control variables between deep learning and optimization theory-based approaches \cite{opt_dnn_01, opt_dnn_04}.
While this partitioning offers a promising direction, these studies generally lack a solid theoretical foundation for effectively integrating the two approaches. A broader application of model-based deep learning is discussed in \cite{model_based_DL}, where certain components of model-based algorithms are replaced with DNNs, simplifying the architecture and reducing the required online training data. However, while this study presents a general methodology for signal processing, control, and communications, it does not specifically focus on the optimization problem formulation in wireless network designs.

In this paper, we introduce an innovative structured and analytical framework for wireless network design that integrates optimization theory with deep learning techniques, addressing the current lack of a theoretical foundation in existing studies. Our methodology begins by constructing the block diagram of the optimization theory-based solution, determining key components tied to optimality conditions and iterative processes. Subsequently, we substitute certain core elements with DNNs at various levels. This unique
blend of optimization theory and deep learning reduces the required training data, minimizes the communication overhead and delay, and enhances the accuracy of the deep learning architectures, while providing adaptivity to the optimization theory-based
algorithms. Furthermore, it significantly improves the interpretability of the solutions by integrating an analytical block diagram into the DNN architecture, rather than relying on a traditional black-box model.


\section{Optimization Theory Based Approach}

The conventional optimization theory based approach starts with the formulation of the optimization problem. This formulation entails specifying decision variables and an objective function to be maximized or minimized. Constraints are also formulated as equations or inequalities that the decision variables must satisfy. Both the objective function and constraints depend on decision variables and system parameters that capture the underlying environmental conditions. For instance, optimizing the transmit power of nodes to maximize network throughput involves not only power levels but also channel gains between nodes. Developing these formulations requires precise mathematical models based on domain expertise, resulting in complex problems involving multiple decision variables across network layers.

The solution strategy for the optimization problem primarily involves deriving optimality conditions and developing iterative solutions, as elaborated below:

\begin{itemize}
    \item \emph{Derivation of optimality conditions}: Optimality conditions define key requirements that must be satisfied at the optimal point by analyzing the objective function and its constraints, including their derivatives at the optimum. Optimization theory-based algorithms use system parameters as inputs, with outputs tied to optimal decision variables. These conditions are applied in two distinct ways within the formal structure of optimization approaches.
    
    \begin{itemize}
        \item \emph{Input set construction:} The optimality conditions linking inputs and outputs establish the relationship between the optimal values of decision variables and the functions of system parameters. Consequently, it is crucial to preprocess the system parameters initially to extract the most pertinent information for the optimization algorithm.
        
        \item \emph{Output set mapping:} The optimality conditions among the outputs define the mathematical relationships governing the optimal values of the decision variables. This means that the algorithm can initially determine the optimal values for a subset of decision variables. Subsequently, the optimal values for the remaining variables can be calculated using mathematical functions derived from the existing ones.
 
    \end{itemize} 

    \item \emph{Development of iterative/heuristic algorithms}: Optimization problems are typically solved using iterative methods or heuristic algorithms designed for manageable complexity while providing solutions close to the optimal. These algorithms involve specific computations tailored to the model as well as mathematical manipulations that remain invariant across different models.

\end{itemize}

The block diagram of the optimization theory-based approach is depicted in Figure \ref{fig_opt}. The system parameters, denoted by $x$, serve as the inputs of the algorithm. These inputs undergo preprocessing using the optimality conditions to extract the most relevant features, represented by $f$. These features are then fed into the iterative algorithm. The iterative algorithm calculates the optimal values for a subset of decision variables, denoted as $y_1$. The remaining decision variables, $y_2$, are derived as functions of the existing ones, utilizing the optimality conditions once more. Together, $y_1$ and $y_2$ constitute the optimal solution. This block diagram outlines fundamental steps for developing optimization theory based algorithms, applicable across various wireless network optimization problems. However, in some cases, certain optimality conditions may not be derived, resulting in missing blocks in the diagram.

\begin{figure*}[!t]
\centering
\includegraphics[width=3.0in]{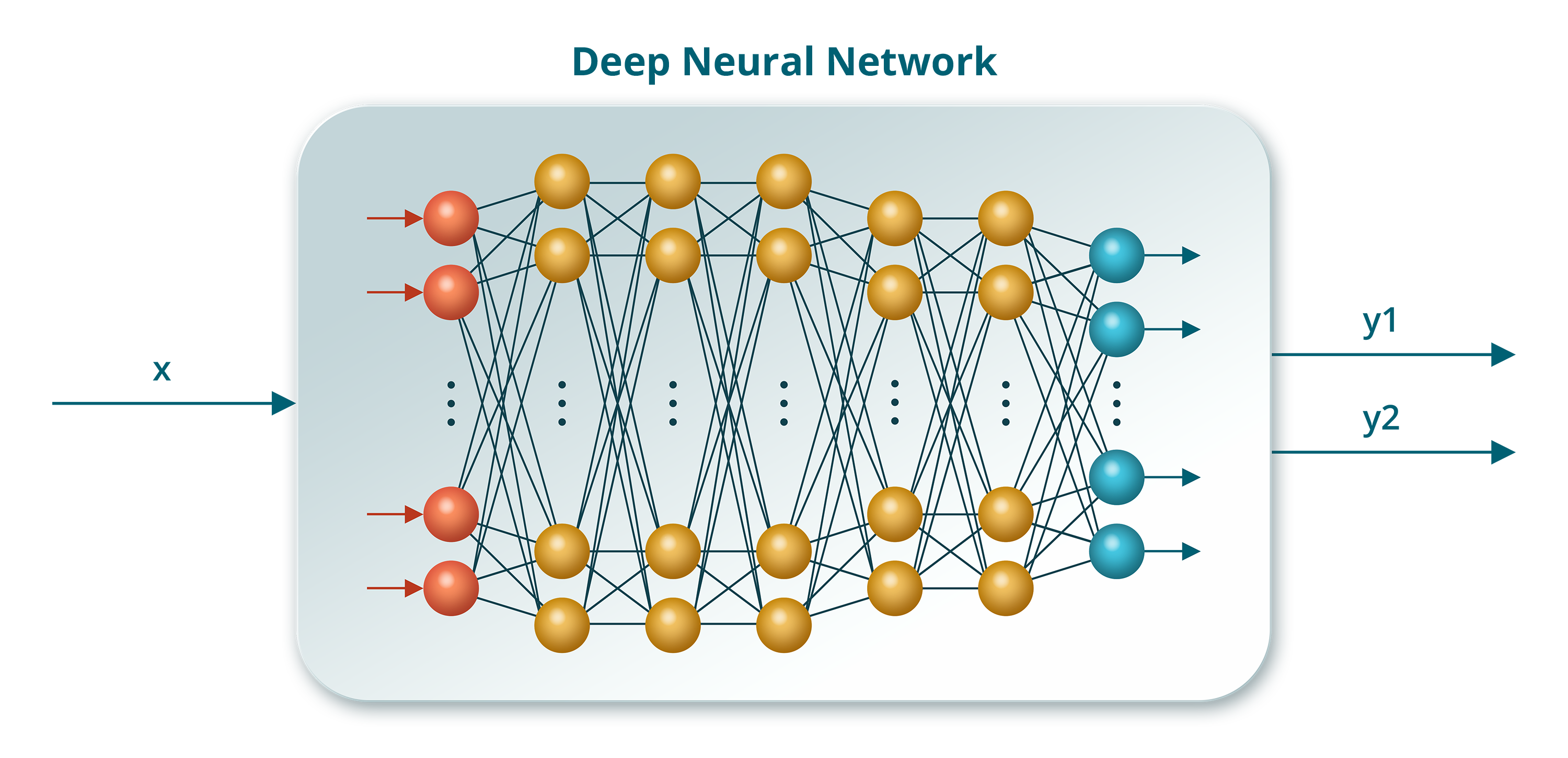}
\caption{Deep learning based data driven approach.}
\label{fig_dnn}
\end{figure*}

\section{Deep Learning based Data Driven Approach}

In the deep learning based data driven approach, the optimization theory-based algorithm is replaced by a DNN. Here, the inputs correspond to the system parameters, and the outputs represent the optimal values of allocated resources, or decision variables, as depicted in Fig. \ref{fig_dnn}. DNNs are universal function approximators for any Borel measurable input-output mapping and are characterized by the weights of the network. Therefore, once DNNs are properly trained, they can be used to obtain the optimal resource allocation every time the system parameters change, in an adaptive manner without relying on any restrictive system model for analytical tractability, at the cost of some inaccuracy due to the approximation inherent in deep learning \cite{Khan22, opt_dnn_03}. The primary challenge in this data-driven approach lies in the training process of DNNs to determine the optimal weights for the mapping. Training DNNs demands considerable amount of time and computing resources, given the need for extensive data. 

Most works in the literature focus on minimizing the amount of online training data required for DNNs. The first set of DNN-based approaches adopts reinforcement learning, where nodes learn through trial and error, using historical values and functions of the system parameters as additional inputs to accelerate the learning process. For instance, in dynamic power allocation \cite{more_inputs}, besides the channel gains, historical values of transmit powers and interference from neighboring nodes are included as inputs to the DNN. However, these inputs are integrated based on the domain knowledge rather than a foundation rooted in optimization theory. 

Another group of DNN-based approaches focuses on reducing the amount of online training data by pretraining them within learning-to-optimize \cite{learn_to_optimize} and knowledge assisted training architectures \cite{digital_twin}. In the learning-to-optimize architecture, the optimal resource allocation is computed for a diverse set of system configurations, representing all possible inputs for the DNN, using an optimization theory-based approach. On the other hand, in the knowledge assisted training architecture, a digital twin of the real network generates representative input-output pairs for the model to pretrain the DNN \cite{digital_twin}. These pretrained DNNs serve as an initial configuration, further fine-tuned in the real environment using actual measured data. All these approaches treat the relationship between system parameters and optimal resource allocation as a black box, without delving into the inner workings of the underlying optimization theory-based methodology.

\begin{figure*}[!t]
\centering
\includegraphics[width=4.0in]{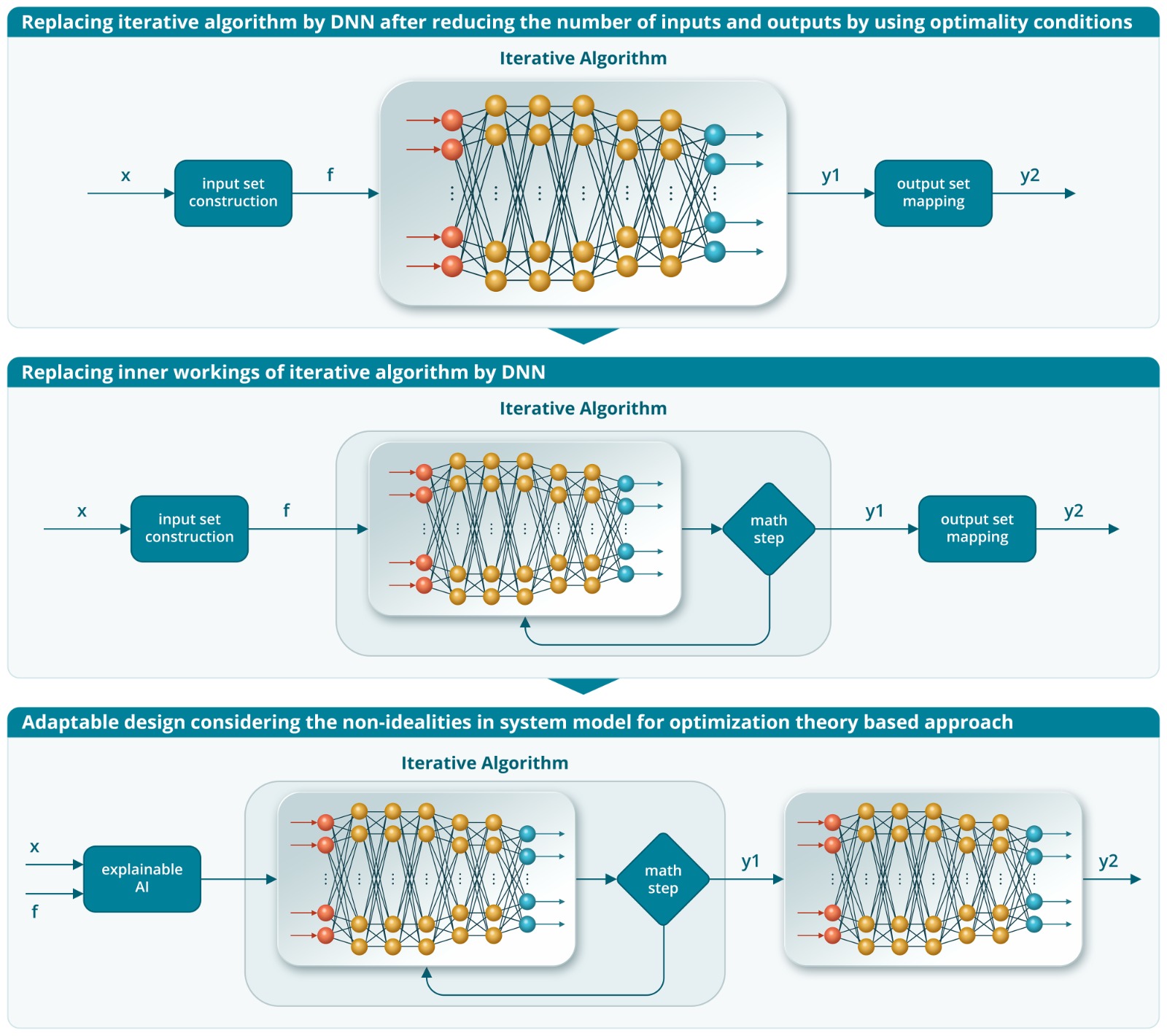}
\label{fig_third_case}
\caption{Optimization theory based deep learning approach.  }
\label{fig_opt_dnn}
\end{figure*}

\section{Optimization Theory Based Deep Learning Approach}


In the optimization theory-based deep learning approach, the optimization theory-based model first identifies key components of the algorithm: input, algorithm, and output building blocks. These components are selectively replaced by DNN architectures, enabling a more structured integration rather than treating the entire algorithm as a black box.

\emph{Input building blocks} correspond to the preprocessing of actual inputs to extract the most valuable features generating outputs. In the optimization theory-based model, system parameters serve as inputs, while the optimal values of decision variables act as outputs. Optimality conditions establish the relationship between the optimal values of the decision variables and functions of the system parameters, denoted by features. In DNN-based architectures, the model's output depends on these features rather than directly on the system parameters, allowing for reduced training time when features are used instead of raw parameters. This approach assumes an accurate theoretical model. In cases where the model accuracy is limited, both system parameters and features can be used as inputs, which is then followed by the analysis of the contribution of each input to the output using feature importance ranking techniques in explainable artificial intelligence (XAI) \cite{XAI}.


\emph{Algorithm building blocks} correspond to generating the building blocks of the optimization theory-based algorithm. This algorithm, whether iterative or heuristic, encompass model specific computations as well as mathematical manipulations that remain invariant across different models. Instead of replacing the entire system as a black box, substituting these building blocks by DNN not only improves the accuracy but also enhances interpretability significantly.


\emph{Output building blocks} establish relationships between subsets of output variables. Optimality conditions are derived to find the mathematical relations among optimal values of decision variables, i.e. outputs. Consequently, only a subset of outputs is obtained through an optimization theory based algorithm, while the remaining outputs are derived mathematically. This approach decreases the training time by reducing the complexity and size of the DNNs, assuming accurate optimization models. If the model lacks accuracy, DNNs can be used to replace these mathematical mappings.

Fig. \ref{fig_opt_dnn} shows different options for the implementation of optimization theory based deep learning approach. In the first implementation, \emph{input building blocks} and \emph{output building blocks} are used to construct the feature set $f$ from the actual inputs, i.e. system parameters denoted by $x$, and establish a mapping in the output set between $y_1$ and $y_2$, respectively. The remaining iterative algorithm is replaced by a DNN. In the second implementation, additionally, the internal operations of the iterative algorithm are considered by using \emph{algorithm building blocks}, and model specific computations in the algorithm are replaced by a DNN. In the third implementation, the approach takes into account the potential inaccuracy of the system model used in the optimization problem. Here, the impact of the features on the final output is evaluated using XAI techniques. Moreover, the mapping between subsets of outputs is replaced by a DNN.

\begin{figure*}[!t]
\centering
\includegraphics[width=6.0in]{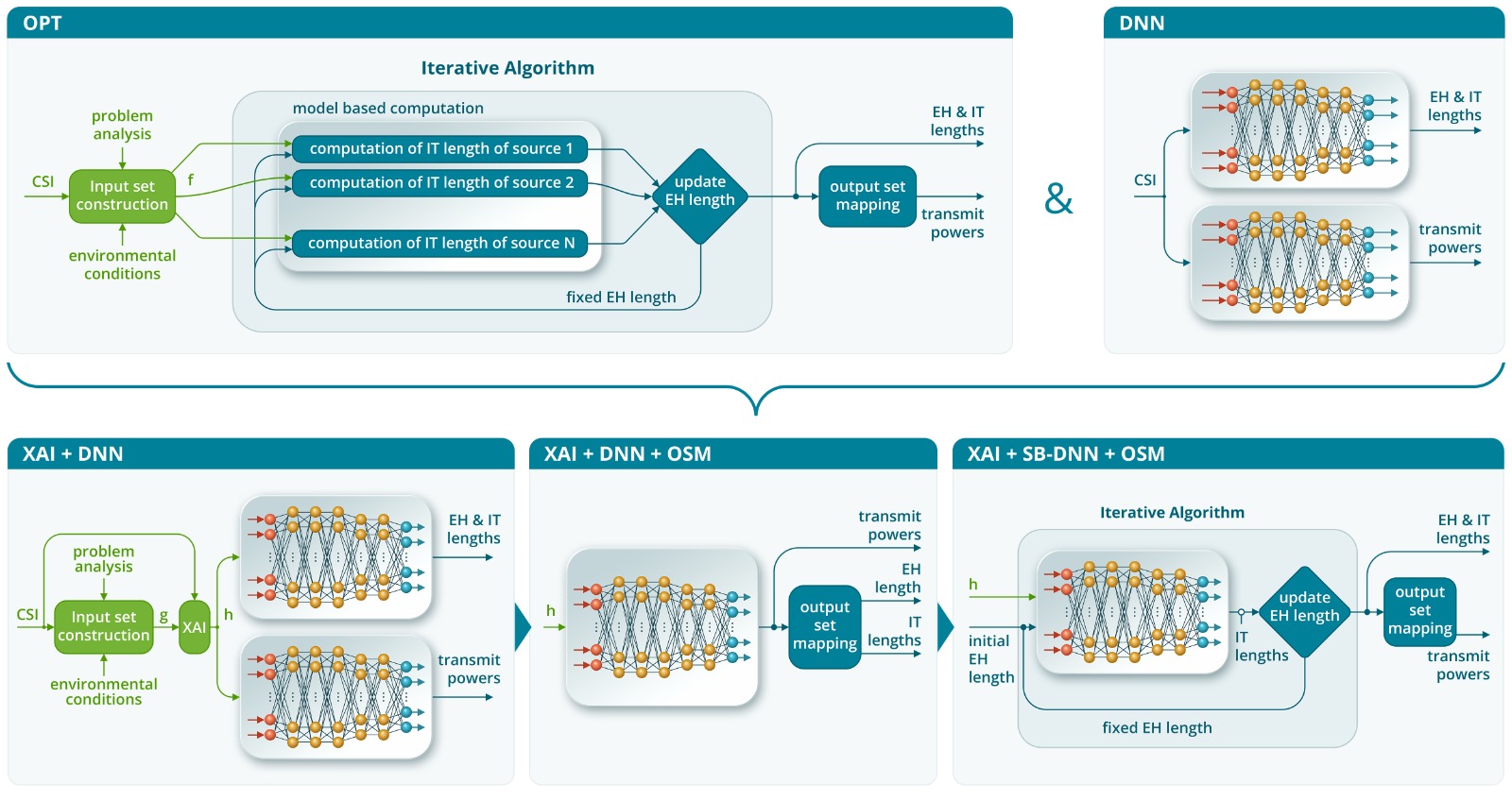}
\label{fig_subdnn} \\
\caption{The block diagrams of OPT, DNN, XAI+DNN, XAI+DNN+OSM and XAI+SB-DNN+OSM, and their relation to each other. 
}
\label{fig_case}
\end{figure*}

\section{Case Study: Minimum Length Scheduling for Wireless Powered Communication Networks}

To illustrate the usage of the optimization theory-based deep learning approach, we consider the minimum length scheduling problem in half-duplex wireless powered communication networks (WPCNs) \cite{opt_3}. The network comprises a single access point (AP) and a designated number of users. During the energy harvesting (EH) phase, the AP broadcasts a downlink signal to the users at a constant transmit power $P_A$, while users harvest energy from this signal with an efficiency rate $\zeta$. Following this, the users sequentially transmit $D$ bits of information to the AP during their allocated information transmission (IT) period, using only the energy harvested during the EH phase. The users' transmit power is assumed to be limited by $P_{max}$ due to hardware constraints. By applying Shannon’s channel capacity formula for additive white Gaussian (AWGN) channels with bandwidth $W$ and noise power spectral density $N_0$, the objective of the problem is to minimize the schedule duration, i.e. the sum of EH and IT durations, while adhering to data demand, energy causality, and maximum transmit power constraints. The system parameters representing environmental conditions, i.e. inputs, are the channel state information (CSI) of the communication links, whereas the decision variables for the optimization problem are transmit powers of the users, EH and IT durations.

\subsection{Implementation of Proposed Algorithms}
\label{proposed_algos}

An optimization theory based iterative algorithm, denoted by OPT, leverages a bi-level transformation of the minimum length scheduling problem based on the convexity of the objective function with respect to the EH length \cite{Onalan20}. In this transformation, the original problem is decomposed into subproblems representing the optimal IT length of each individual source for a fixed EH length. For each source, denote the IT length for transmitting at maximum power as $\beta$ and the required EH duration for transmitting at maximum power as $\theta$. For a given EH duration, if $\theta$ is less than the available EH time, the optimal IT duration equals $\beta$; otherwise, the optimal IT duration is expressed as a function of parameters derived from the optimality conditions of the single-source problem, represented by $\alpha$ and $\gamma$. Consequently, the feature vector $f = [\alpha, \gamma, \beta, \theta]$ is generated for each user through the Input Set Construction (ISC) step to calculate the individual IT durations for a given EH period. These subproblems are coordinated by a master problem, which iteratively searches for the optimal EH duration using the bisection method. The result of this search provides the optimal EH and IT durations, from which the optimal transmit powers are then computed.

The deep learning-based data-driven approach, denoted by DNN, employs two DNN architectures: one for EH and IT lengths, and the other for transmit powers. The inputs for both DNNs consist of uplink and downlink CSI of all sources, resulting in an input size twice the number of sources. The outputs of one DNN are the power allocations of each source, and the outputs of the other are the EH time for all sources and IT time allocations of each source. Both DNNs are multi-input, multi-output, feed-forward, fully connected, and consist of five hidden layers with the size of  $8$, $8$, $8$, $4$, and $4$ times the number of sources. These hyperparameters are determined using grid search, selecting the architecture with minimal training and validation loss to prevent overfitting. Rectified Linear Unit (ReLU) activation functions are employed in each hidden layer. The power outputs are capped by a maximum transmit power constraint, enforced using the sigmoid function. The DNNs are trained using a joint loss function, a weighted sum of mean square error (MSE) and data outage amount. For MSE calculation, the truth labels are obtained by OPT. Data outage refers to the unmatched amount of data demand when sources transmit with the predicted powers during predicted times. 


The DNN architecture expanded to include additional features at the input, with a subsequent selection of the most impactful inputs using XAI techniques, is denoted by XAI+DNN. This architecture incorporates the explainable AI block in the third implementation of Fig. \ref{fig_opt_dnn}, due to the absence of a direct mapping from the actual inputs to the feature set. ISC outputs the vector $g=[\alpha,\gamma]$. $\beta$ and $\theta$ values are omitted since the maximum transmit power is controlled at the output layer of the DNN.  Additional input parameters, $\alpha,\gamma$, derived from the optimality conditions of the problem, can improve the accuracy of the aforementioned DNN architecture. The most impactful features, denoted by $h$, among $\alpha,\gamma$ and CSI are then selected based on the mutual information (MI) between all possible inputs and the schedule length in XAI. 

The XAI+DNN architecture extended by the output set mapping (OSM) based on the optimality conditions of the problem is called XAI+DNN+OSM. This architecture incorporates the output set mapping block in the first and second implementation of Fig. \ref{fig_opt_dnn}, based on the derivation of the optimal IT and EH durations for given transmit powers in \cite{Onalan22}. Instead of using one DNN for power and another for time allocations, the output layer of the DNN  predicting power allocations can be directly mapped to the corresponding optimum time allocation. This mapping simplifies the DNN outputs and ensures a feasible solution. 

 \begin{figure*}[!t]
\centering
\includegraphics[width=6.0in]{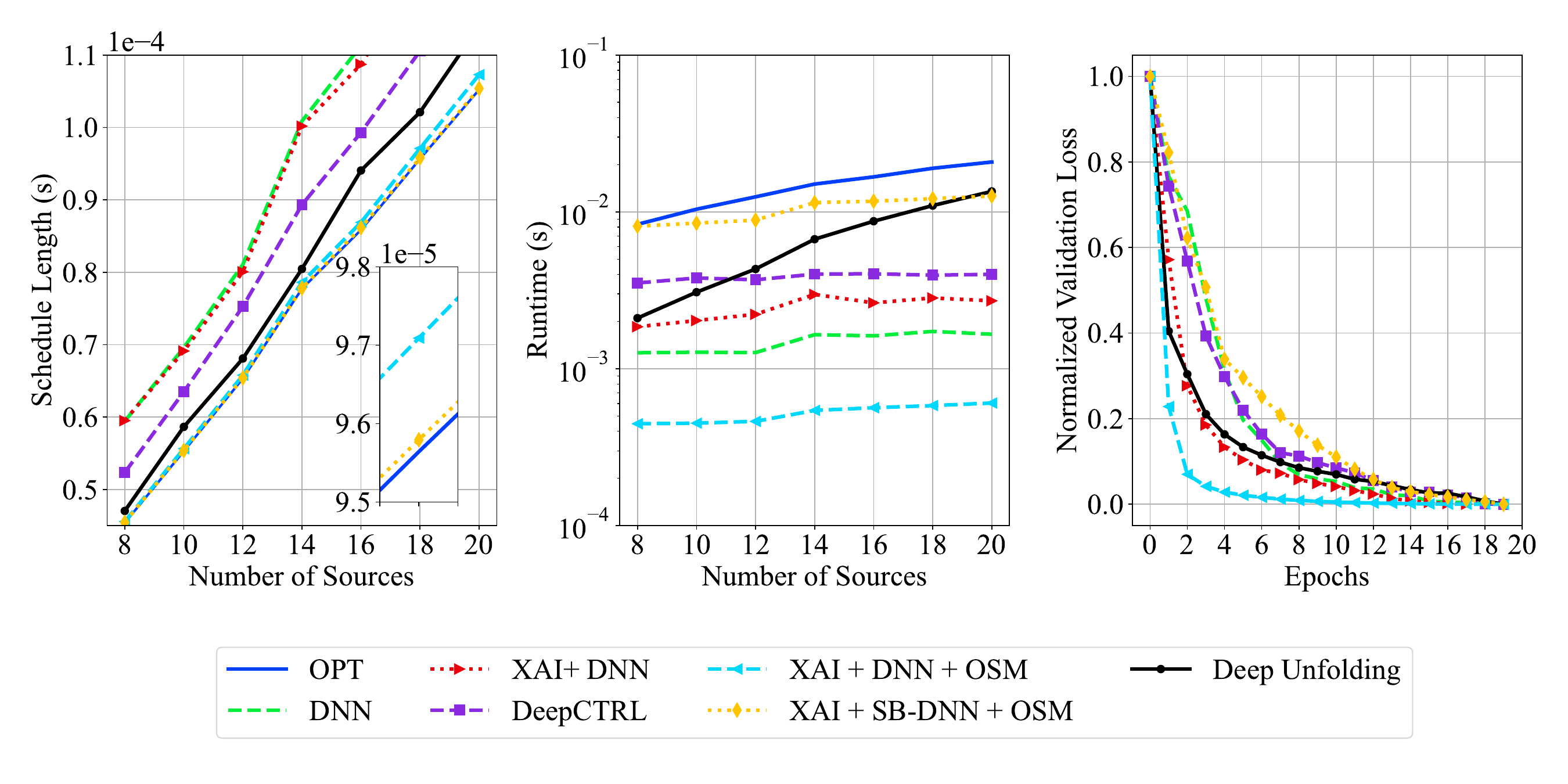}
\caption{Schedule length, runtime and normalized validation loss of the algorithms.}
\label{fig:schedule_runtime_result}
\end{figure*}


    

The DNN that replaces the inner workings of the iterative algorithm in addition to applying XAI at the input and OSM at the output is denoted by XAI+SB-DNN+OSM. This architecture replaces the model specific computations in the algorithm building blocks by DNN, as shown in the second and third implementation of Fig. \ref{fig_opt_dnn}. OPT derives individual IT lengths by employing numerical analysis techniques to solve complex mathematical equations for a given EH length in each sub-problem. Here, a DNN replaces these derivations. The DNN uses a feed-forward, fully connected architecture with five hidden layers to estimate the individual IT times for a fixed EH length. The inputs of the DNN are the vector $h$ and the fixed EH length; whereas the outputs are the IT lengths of each source. The loss function is expressed in terms of total unsatisfied data demand when sources transmit with the predicted powers during predicted times, i.e., data outage amount. Then, the optimal EH length is again searched by the bisection method, whose outputs are the EH and IT lengths and are mapped to transmit powers as in OPT. 

Fig. \ref{fig_case} shows the block diagrams of the implementation of these proposed algorithms and their relation to each other. 




\subsection{Performance Evaluation}
For performance comparison, we selected DeepCTRL and deep unfolding, along with the optimization theory based iterative algorithm OPT and the deep learning-based data-driven approach DNN, as described in Section \ref{proposed_algos}, to serve as benchmarks. DeepCTRL couples data representations with rules of physical worlds by employing separate data and rule encoders, along with a decision encoder to merge their outputs \cite{deepctrl}. 
DeepCTRL is adapted to the minimum length scheduling problem in WPCNs by combining a problem-specific rule encoder, trained to prevent the data outage, with a data encoder, trained to minimize the MSE loss. This sets DeepCTRL apart from the XAI+DNN, XAI+DNN+OSM, and XAI+SB-DNN+OSM approaches, which incorporate optimization theory-based components in place of the rule encoder to enhance accuracy. On the other hand, deep unfolding is based on unfolding the repeated iterations into T blocks boosted by DNNs \cite{deep-unfold}, where T is typically much smaller than the number of iterations required by optimization theory-based approaches. Deep unfolding is adapted by replacing the update state of the iterative algorithm by a DNN for T blocks, distinguishing it from the XAI+SB-DNN+OSM approach, which replaces model-based computations while preserving OPT iterations. 
We set T as half of the number of sources plus one. We apply end-to-end supervised training. 



The dataset is constructed by uniformly distributing information sources within a circular area of a radius of $1$ m where the access point (AP) is located at the center. CSI is determined by considering both large and small-scale statistics. The large-scale fading is modeled by $PL(d) = PL(d_0) + 10 \upsilon log(d/d_0) + Z$, where $d$ is the distance between the user and the AP, $PL(d)$ is the path loss at distance $d$ in decibels, $PL(d_0)$ is the path loss at the reference distance $d_0 = 1$ m, $\upsilon=2$ is the path loss exponent and Z is a Gaussian random variable with zero mean and standard deviation $\sigma_z=2$dB. The small-scale fading is modeled as a Rayleigh distributed random variable with scale parameter $\Omega$ set to the mean power level determined by $PL(d)$. 
The following environmental simulation parameters are considered:   $P_{max}=10$ mW,  $P_{A}=2$ W, $N_0=-110$ dBm, $W=1$ MHz, $\zeta= 50$\% , $D= 50$ bits.

All DNN architectures are implemented in Pytorch and trained by Adam optimizer with a learning rate $10^{-4}$, weight decay $10^{-3}$, and batch size $32$. Training is limited to a maximum of $100$ epochs, with early stopping applied if the validation loss improves by less than $5$\% over $5$ consecutive epochs. Weights and biases are initialized with a zero-mean uniform random distribution, where the variance is scaled according to the input size.
In all deep learning approaches, reducing the input size by XAI techniques creates a tradeoff between accuracy and complexity. An input size of three times the number of sources yields the best performance and is used in all XAI-based methods. The generated data is divided into train, validation, and test sets with sizes of $100,000; 1,000;$ and $1,000,$  respectively. During training, we monitor training and validation losses to prevent overfitting, with DNN architectures and the weight decay parameter in the Adam optimizer further helping to mitigate this risk. Performance evaluation is conducted by averaging results over the test set. The confidence intervals around these averages remain consistent across different algorithms and are omitted for simplicity. The software for the implementation of the algorithms is available at \cite{code}.


Fig.~\ref{fig:schedule_runtime_result} shows the trade-off between the schedule length and the runtime of the algorithms, highlighting their scalability for different numbers of sources. OPT reaches the minimum schedule length with the highest runtime, as expected. The runtime of solving sub-problems, e.g., via fixed point iterations, rapidly increases with the number of sources. XAI+SB-DNN+OSM performs very close to OPT with a runtime improvement due to the usage of DNN in the solution of the sub-problems. The difference in the runtime of OPT and XAI+SB-DNN+OSM increases as the network size increases, demonstrating superior scalability in XAI+SB-DNN+OSM. In DNN and XAI+DNN, the outputs may not satisfy the data demands due to the prediction errors despite the loss function enforcing the minimization of data outage amount. In such cases, to ensure a feasible solution, scheduling is repeated for sources with remaining information to send. This repetition causes a longer schedule length than the optimum. XAI+DNN utilizes additional features compared to DNN, resulting in lower schedule length but higher runtime than DNN.  DeepCTRL provides higher accuracy since it also benefits from the rule encoder. The disadvantage of the additional rule encoder is additional runtime compared to DNN. XAI+DNN+OSM provides close to optimum solutions with up to 2\% gap as OSM simplifies the learning process during training and ensures a feasible solution. XAI+DNN+OSM also results with the lowest runtime since it uses only one DNN architecture, thanks to OSM. Deep unfolding provides competitive results in terms of schedule length; whereas its runtime increases as T increases with number of sources. In the trade-off between runtime scalability and optimality, XAI+DNN+OSM offers the best scalability with the lowest runtime, whereas XAI+SB-DNN+OSM delivers the highest accuracy with slightly less scalability in terms of runtime.

Fig.~\ref{fig:schedule_runtime_result} further illustrates the normalized validation loss over the training epochs to compare the convergence rates of different algorithms. The validation loss is normalized using max-min normalization, as each algorithm employs distinct validation loss functions due to their varied architectures and parameter predictions. Generally, algorithms converge more rapidly when their architectures are simpler or the prediction problems are less complex. A comparison between XAI+DNN and DNN reveals that incorporating additional features and subsequently selecting the most impactful ones with XAI accelerates convergence. XAI+DNN+OSM demonstrates the fastest convergence, highlighting the effectiveness of extending the architecture with the output set mapping based on optimality conditions. In contrast, XAI+SB-DNN+OSM requires consideration of the initial EH length, introducing additional complexity and resulting in slower convergence compared to the other algorithms.

\section{Conclusion}
 In this paper, we introduce an innovative method that integrates optimization theory and deep learning techniques for wireless network design. The methodology initiates by formulating the optimization problem, incorporating feature extraction through optimality conditions in the input, mapping a subset of outputs to the rest, and utilizing iterative solutions rooted in optimization theory. In the subsequent deep learning phase, specific components of the optimization theory-based approach are substituted with deep neural networks. Through extensive simulations, we demonstrate that utilizing an optimization theory-based architecture for implementing deep neural networks leads to better runtime scalability, improved accuracy, and a better convergence rate, while minimizing communication overhead and delay. The addition of optimization theory-based features enhances accuracy and convergence rate, albeit with a marginal increase in runtime. Furthermore, employing optimization theory-based output set mapping significantly enhances runtime, accuracy, and convergence rate. Substituting inner blocks of the optimization theory-based iterative approach with deep neural networks significantly boosts accuracy, with a moderate increase in runtime that remains lower than that of a pure optimization theory-based approach. In the future, we plan to apply the proposed methodology to complex optimization problems formulated in various contexts, including intelligent reflecting surfaces (IRS)-assisted wireless networks, massive multiple-input multiple-output (MIMO) wireless networks, and wireless networked control systems. This will involve first constructing a block diagram of the optimization theory-based solution and then selectively replacing key blocks with DNNs, followed by a comprehensive performance evaluation.


%



\ifCLASSOPTIONcaptionsoff
  \newpage
\fi



%

\bibliographystyle{IEEEtran}
\bibliography{dl_bib}

\vskip -2\baselineskip plus -1fil
\begin{IEEEbiographynophoto}{Sinem Coleri}
received her Ph.D. in electrical engineering and computer sciences from University of California, Berkeley, in 2005. She is currently Professor at Koc University.
\end{IEEEbiographynophoto}

\vskip -3\baselineskip plus -1fil

\begin{IEEEbiographynophoto}{Aysun Gurur Onalan}
received her Ph.D. in electrical and electronics engineering from Koc University in 2023. She is currently research team lead at Lifemote Networks.
\end{IEEEbiographynophoto}
\vskip -3\baselineskip plus -1fil

\begin{IEEEbiographynophoto}{Marco Di Renzo}
received his Ph.D. degree in electrical engineering from the University of L’Aquila, Italy, in 2007. He is currently a CNRS Research Director (Professor) at CentraleSupélec - Paris-Saclay University, France.
\end{IEEEbiographynophoto}






\end{document}